\documentclass[10pt,twocolumn,letterpaper]{article}

\usepackage{iccv}
\usepackage{times}
\usepackage{epsfig}
\usepackage{graphicx}
\usepackage{amsmath}
\usepackage{amssymb}

\usepackage{booktabs}
\usepackage{multirow}
\usepackage{multicol}
\usepackage{kotex}
\usepackage[breaklinks=true,bookmarks=false]{hyperref}

\iccvfinalcopy 

\ificcvfinal\fi

\usepackage{xcolor}

\begin{document}

\title{Fine-Grained Attention for Weakly Supervised Object Localization}

\author{Junghyo Sohn$^{1}$ \quad Eunjin Jeon$^{2}$ \quad Wonsik Jung$^{2}$ \quad Eunsong Kang$^{2}$ \quad Heung-Il Suk$^{1,2,}$\thanks{Corresponding author.}\\
$^1$Department of Artificial Intelligence, Korea University, Korea\\
$^2$Department of Brain and Cognitive Engineering, Korea University, Korea\\
{\tt\small \{jhsohn0633, eunjinjeon, ssikjeong1, eunsong1210, hisuk\} @korea.ac.kr}}

\maketitle
\ificcvfinal\thispagestyle{empty}\fi

\begin{abstract}
Although recent advances in deep learning accelerated an improvement in a weakly supervised object localization (WSOL) task, there are still challenges to identify the entire body of an object, rather than only discriminative parts. In this paper, we propose a novel residual fine-grained attention (RFGA) module that autonomously excites the less activated regions of an object by utilizing information distributed over channels and locations within feature maps in combination with a residual operation. To be specific, we devise a series of mechanisms of triple-view attention representation, attention expansion, and feature calibration. Unlike other attention-based WSOL methods that learn a coarse attention map, having the same values across elements in feature maps, our proposed RFGA learns fine-grained values in an attention map by assigning different attention values for each of the elements. We validated the superiority of our proposed RFGA module by comparing it with the recent methods in the literature over three datasets. Further, we analyzed the effect of each mechanism in our RFGA and visualized attention maps to get insights.
\end{abstract}

\begin{figure}[t]
\begin{center}
   \includegraphics[width=1.0\linewidth]{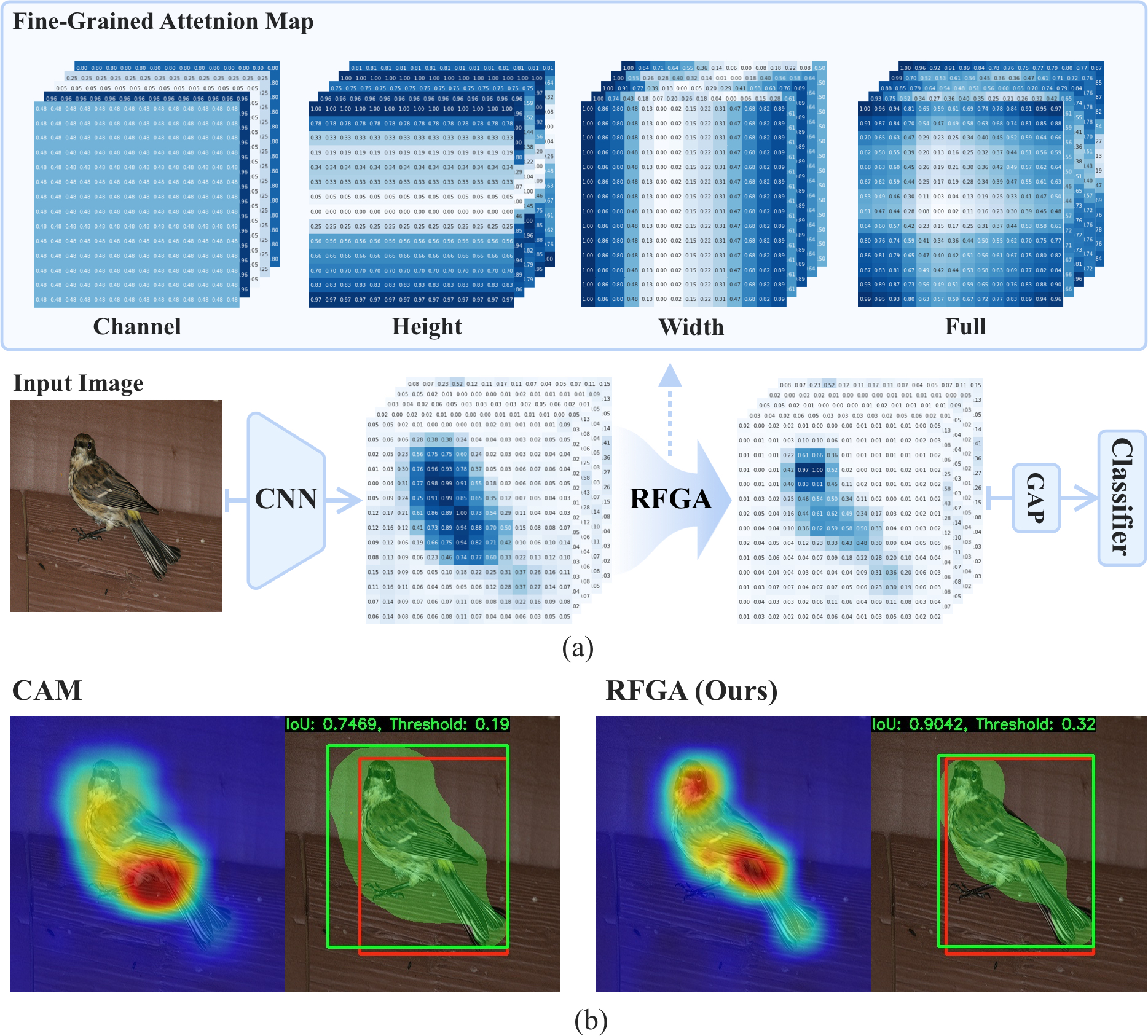}
\end{center}
   \caption{
   (a) Overview of our RFGA method that generates fine-grained attended maps to be efficient for WSOL by considering triple-view attentions (channel, height, and width before a classifier. The full attention map is generated by an outer sum of triple-view attentions.
   (b) Comparison of CAM \cite{Zhou_2016_CVPR} and our RFGA with respect to an activation map (left) and a localization (right). For localization, red and green boxes denote the ground-truth and predicted bounding boxes, respectively. In addition, green-masked region indicates the activation map after applying a threshold.
   }
\label{fig:onecol}
\end{figure}

\section{Introduction}
During the last decade, researchers have developed various forms of deep learning-based models and achieved remarkable performance in object localization of inferring the bounding box of objects in natural images \cite{felzenszwalb2009object,girshick2014rich,ren2016faster}. However, from the learning and data efficiency perspectives, the major limitation of those works is the use of a fully-labeled dataset for supervision. Undoubtedly, it is labor-intensive and time-consuming to make such a fully-labeled dataset, thus causing their applicability limited in practice.  

Meanwhile, Weakly Supervised Object Localization (WSOL) methods, which employ only class labels but no information of the targeted bounding box of objects \cite{Zhou_2016_CVPR,wei2017object,zhu2017soft,singh2017hide,zhang2018adversarial,zhang2018self,choe2019attention, xue2019danet,mai2020erasing,babar2021look}, has attracted by showing their great potential for the same task, being trained in a more data-efficient manner. The main idea of WSOL methods is to detect the class-discriminative regions via an object recognition task and to utilize those regions for the localization of the identified object.
For example, Class Activation Map (CAM) \cite{Zhou_2016_CVPR} that estimates the class-specific discriminative regions based on the inferred class scores is one of the representative methods in WSOL. In the meantime, various studies \cite{Zhou_2016_CVPR,zhu2017soft,wei2017object,singh2017hide,zhang2018adversarial,zhang2018self,choe2019attention,xue2019danet,mai2020erasing,babar2021look,choe2020evaluating} addressed that CAM-based methods is not capable of capturing the overall object regions in a finer way, because it focuses on only the class-discriminative regions, disregarding non-discriminative regions. Hence, many of the output bounding boxes are not tight enough to the target object by resulting in either over-sized or under-sized. There have been efforts to tackle these challenges via diverse network architectures or learning strategies \cite{wei2016stc, wei2017object, singh2017hide,zhu2017soft, kim2017two, zhang2018adversarial,choe2019attention,mai2020erasing,yun2019cutmix, choe2019attention, mai2020erasing, babar2021look}.

In principle, those methods devised different kinds of mechanisms to mitigate the major problem of focusing on only the discriminative regions in localization, by intentionally corrupting (\ie, erasing) an input image \cite{yun2019cutmix, singh2017hide, wei2017object} or a feature \cite{zhang2018adversarial, mai2020erasing} or generating an attention map. For the image corruption methods, two different strategies were exploited, namely, random corruption and network-guided corruption. First, the random corruption approach removes a small patch within an image at random and uses the corrupted image to learn richer feature representations \cite{singh2017hide, yun2019cutmix}. This helps the trained network to discover diverse discriminative representations, thus to detect more object-related regions. The network-guided corruption approach adaptively corrupts by dropping out the most discriminative regions based on the integrated activation maps \cite{wei2017object, zhang2018adversarial,choe2019attention,mai2020erasing}. As for the attention-based methods \cite{choe2019attention, zhang2018self}, they use a specially designed module to generate an attention map, based on which the most discriminative regions are hidden to capture the integral extent of an object.

While those methods helped improve the performance, they have limitations and issues that should be considered further. First, the random-corruption approach \cite{singh2017hide, yun2019cutmix} potentially disrupts the network learning due to unexpected information loss \cite{zhang2018adversarial,choe2019attention}. For example, if the object-characteristic parts were removed from an input image, a network is enforced to discover other parts from the remaining regions. Obviously, when there exists no further discriminative region, it would be trained in a wrong way. Second, the network-guided corruption approach introduces additional hyperparameters to determine the most discriminative regions and their sizes in activation maps. 
Third, the attention-based methods \cite{wei2017object,zhang2018adversarial,choe2019attention,mai2020erasing} mostly exploit the coarse information in the form of channel or spatial attention and apply the same attention values to units in feature maps.

In this paper, we propose a novel fine-grained attention method that efficiently and accurately localizes object-related regions in an image. Specifically, we propose a new mechanism to generate a fine-level attention map that allows to utilize a series of information distributed over channels and locations within feature maps. The fine-level attention map is in the same size of as the input feature maps to the attention-based module, thus the attention is assigned for each of the units across feature maps and channels. Compared to the corruption-based approaches, our proposed method doesn't need to mask patches in an image and doesn't have additional hyperparameters for most discriminative regions selection. 

The main contributions of our work are three-fold:
\begin{itemize}
    \item We propose a novel mechanism to represent a fine-grained attention that allows us to utilize feature representations globally in high resolution, thus to localize an object accurately.
    \item In combination with a residual connection, our attention module autonomously concentrates on the less activated regions. Accordingly, it is more likely to focus on other informative regions of an object in an image.
    \item In the experiments, our proposed method, Residual Fine-Grained Attention (RFGA), achieved the state-of-the-art object localization performances in the metrics of mIOU and MaxBoxAcc \cite{choe2020evaluating} on three datasets, \ie, CUB-200-2011 \cite{wah2011caltech}, FGVC Aircraft \cite{maji2013fine}, and Stanford Dogs \cite{KhoslaYaoJayadevaprakashFeiFei_FGVC2011}.
\end{itemize}

\section{Related Work}
\textbf{Weakly supervised object localization.}
WSOL can be mainly categorized into two approaches depending on the selection method of erasing (\ie, corrupting) regions: (1) random corruption \cite{singh2017hide,yun2019cutmix} and (2) network-guided currption methods \cite{zhang2018adversarial,zhang2018self,choe2019attention,mai2020erasing}. With regard to the random corruption method, Singh and Lee \cite{singh2017hide} devised Hide-and-Seek (HaS), which randomly drops the patches of input images in order to encourage the network to find other relevant regions rather than only focus on the most discriminative parts of an object. Yun \etal~\cite{yun2019cutmix} introduced CutMix where the randomly erasing (\ie, cutting) patches are filled with patches of another class and the corresponding labels are also mixed. Though these methods have been considered as an efficient data augmentation method due to their no requirement of parameters, the random corruption can negatively affect localization performance due to its brute-force elimination of the input images \cite{babar2021look,zhang2018adversarial,choe2019attention}.

For the network-guided corruption methods \cite{zhang2018adversarial,choe2019attention,zhang2018self,mai2020erasing}, the most discriminative regions of the original image or feature map are dropped with a threshold (\ie, drop rate). Zhang \etal ~\cite{zhang2018adversarial} proposed an Adversarial Complementary Learning (ACoL) to find the complementary regions through an adversarial learning between two parallel-classifiers; one to erase discriminative regions and the other to learn other discriminative regions except for the erased regions. Similar to ACoL, Choe \etal~\cite{choe2019attention} introduced an Attention-based Dropout Layer (ADL), which generates a drop mask and an importance map by utilizing the self-attention mechanism and then randomly selects one of them for thresholded feature maps. In addition to these methods, \cite{zhang2018adversarial,choe2019attention,zhang2018self,mai2020erasing} also exploited a self-attention mechanism to identify discriminative regions. However, they all require a drop rate as a criterion of the masking. In these regards, our proposed RFGA is capable of discovering from the less and to the more discriminative regions by using a novel self-attention module, without setting up the drop rate.

\textbf{Attention based Deep Neural Networks.}
Attention mechanisms have widely used to enhance the representational power of features for their tasks. Among various attention mechanisms \cite{yue2018compact, gao2020channel, wang2020axial, zheng2019looking, gao2020kronecker, huang2019ccnet, wang2018non, zhao2020exploring, cao2019gcnet}, here, we focus on a context fusion based mechanism
\cite{hu2018squeeze, woo2018cbam, wang2020eca, lee2019srm, zhuang2020learning, liu2020improving, kim2020spatially, lee2019srm, hu2018gather, kim2020learning, yang2021sa} that strengthens the feature maps to be more meaningful by aggregating information from every pixel. For instance, Hu \etal~\cite{hu2018squeeze} proposed a Squeeze-and-Excitation Network (SENet), a highly simple and efficient gating mechanism to consider the channel-wise relationships among feature maps of the basic architectures. Likewise, Woo \etal~\cite{woo2018cbam} devised a Convolutional Block Attention Module (CBAM) that sequentially combines two separate attention maps for channel and spatial. Different from SENet \cite{hu2018squeeze}, CBAM \cite{woo2018cbam} additionally considered spatial attention which involves ``where'' to focus. Moreover, to alleviate a limitation of SENet \cite{hu2018squeeze} that utilizes fully-connected layers in order for the reduction of the computational cost at the expense of the association between channel and weight, Wang \etal~\cite{wang2020eca} introduced an Efficient Channel Attention Network (ECA-Net) \cite{wang2020eca} that deploys a 1D convolutional layer to obtain cross-channel attention, while maintaining lower model complexity.

However, since \cite{hu2018squeeze,woo2018cbam,wang2020eca} emphasized meaningful features by multiplying the same attention values, where the different information corresponding to spatial (\ie, height and weight) or channel dimensions might be ignored, they can be unsuitable for WSOL where the fine location information is demanded. Meanwhile, our RFGA generates a detailed attention map that has different attention values across all regions by inferring the intersection of triple-view (\ie, height, weight, and channel) attentions.

\begin{figure*}[t]
    \centering
    \includegraphics[width=1\linewidth]{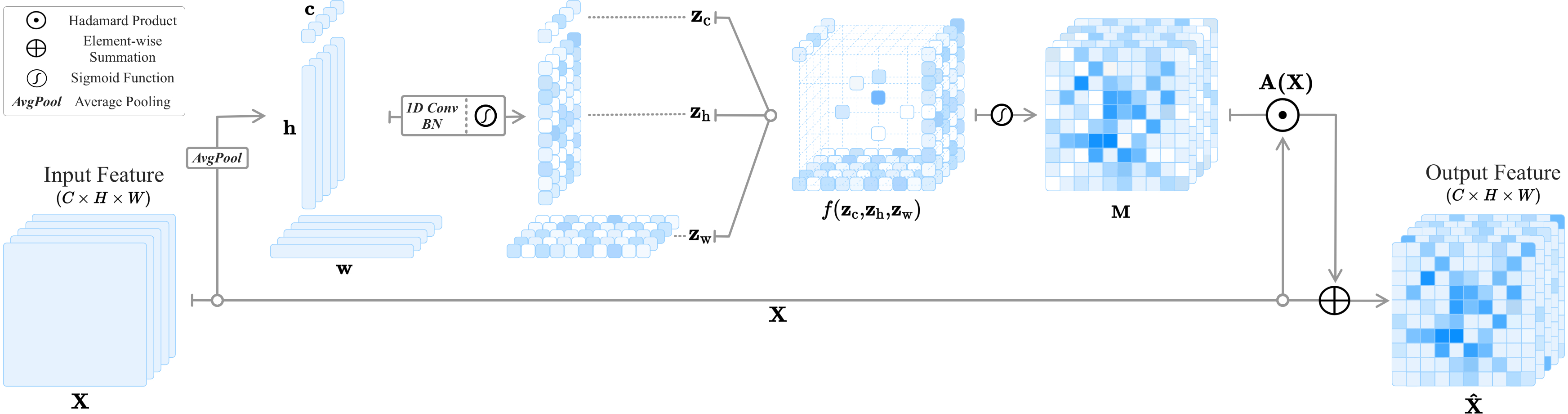}
    \caption{An illustration of Residual Fine-Grained Attention (RFGA) module. An input feature $\mathbf{X}$ is processed by triple-view attentions transformed from three kinds of pooled features, $\mathbf{h}, \mathbf{w}$, and $\mathbf{c}$, which are then fed into a expansion function $f$. The generated fine-grained attention map is combined with the input feature, which is referred to $\mathbf{A}(\mathbf{X})$. Through a residual connection between $\mathbf{A}(\mathbf{X})$ and $\mathbf{X}$, we obtain $\hat{\mathbf{X}}$ and feed it into a classifier.}
    \label{fig:overview}
\end{figure*}

\section{Residual Fine-Grained Attention}
In this section, we present the details of our proposed residual fine-grained attention (RFGA) module. The RFGA is applied on the output feature maps before feeding into a classifier (Fig. \ref{fig:onecol}) to induce the model to learn the entire region of an object. Hereafter, we regard the output feature maps as a feature tensor (3D) without loss of generality.

Our RFGA generates a self-attention tensor, which is generated from three types of view-dependent attention maps by projecting the input feature tensor into channel, width, and height dimensions, respectively. On the contrary to this, the existing works \cite{hu2018squeeze, wang2020eca} primarily consider channel-wise attention by ignoring the spatial characteristics distributed over the different maps in a feature tensor. The RFGA-generated attention tensor presents a fine-grained characteristic in the sense of assigning different attention values for each of the elements in a tensor. Notably, the residual connection in RFGA leads the attention tensor to focus on less discriminative areas of an object as well. In these regards, the final output feature tensor presents an enriched representation resulting in better object localization output, even relatively less discriminative feature for classification. The overall architecture of the proposed RFGA is illustrated in Fig. \ref{fig:overview} and the detailed descriptions are given below.

\subsection{Triple-view Attentions}
Let $\mathbf{X}\in\mathbb{R}^{C\times H\times W}$ be an input feature tensor, where $C$, $H$, and $W$ denote the dimension of the channel, height, and width, respectively. To condense the global distribution of an input feature tensor $\mathbf{X}$ in the triple views, we apply an average pooling in each dimension of the tensor, \ie, channel, height, and width as follows:
\begin{gather}
    \mathbf{c} = \text{AvgPool}_\text{wh}(\mathbf{X})\label{eq:avgpool_c}\\
    \mathbf{h} = \text{AvgPool}_\text{w}(\mathbf{X})\label{eq:avgpool_h}\\
    \mathbf{w} = \text{AvgPool}_\text{h}(\mathbf{X})\label{eq:avgpool_w}
\end{gather}
where $\text{AvgPool}_{\{\cdot\}}$ is an average pooling operator with respect to the dimensions of $\{\cdot\}$.

After computations in Eq. \eqref{eq:avgpool_c}-\eqref{eq:avgpool_w}, the three pooled features of $\mathbf{c}\in\mathbb{R}^{C\times1\times1}$, $\mathbf{h}\in\mathbb{R}^{C\times H\times 1}$, and $\mathbf{w}\in\mathbb{R}^{C\times 1\times W}$ can be regarded as a summary of the extracted features in $\mathbf{X}$ from different viewpoints. Surely, the three views carry different information distributed in the input feature tensor $\mathbf{X}$. That is, $\mathbf{c}$ captures which feature representations are highly activated, and $\mathbf{h}$ and $\mathbf{w}$ reflect, respectively, where the discriminative features distributed vertically and horizontally across channels.

Subsequently, in order to utilize their local interaction among units in each pooled feature \cite{wang2020eca}, we apply a 1D convolution with a kernel size of $k$ and zero-padding without biases, thus keeping their dimensionality. Then, a batch normalization \cite{ioffe2015batch} and a non-linear activation function are applied as follows:
\begin{gather}
    \mathbf{z}_\text{h} = \sigma(\text{BN}(\mathbf{W}_\text{h}(\mathbf{h})))\label{eq:z_h}\\
    \mathbf{z}_\text{w} = \sigma(\text{BN}(\mathbf{W}_\text{w}(\mathbf{w})))\label{eq:z_w}\\
    \mathbf{z}_\text{c} = \sigma(\text{BN}(\mathbf{W}_\text{c}(\mathbf{c})))\label{eq:z_c}
\end{gather}
where $\sigma(\cdot)$ is a sigmoid function and $\mathbf{W}_{\{\cdot\}}$ indicates the 1D convolutional layer for the respective pooled features. Here, $\mathbf{z}_\text{h}\in\mathbb{R}^{C\times H\times 1}$, $\mathbf{z}_\text{w}\in\mathbb{R}^{C\times 1\times W}$, and $\mathbf{z}_\text{c}\in\mathbb{R}^{C\times 1\times 1}$ corresponds to the resulting triple-view attentions.

\subsection{Attentions Expansion}
\label{sec:attention}
We propose to expand the triple-view attentions of $\mathbf{z}_\text{h}$, $\mathbf{z}_\text{w}$, and $\mathbf{z}_\text{c}$ in the size of an input feature tensor $\mathbf{X}$, by which it is beneficial to reflect the attention information back into the input feature tensor $\mathbf{X}$ in a fine-grained manner. Therefore, we create an attention map $\mathbf{M}\in\mathbb{R}^{C\times H\times W}$ of the same size of the input feature map $\mathbf{X}$ by means of an expansion function $f$ defined by an outer sum as follows:
\begin{equation}\label{eq:intersection}
    \mathbf{M} = \sigma(f(\mathbf{z}_\text{h}, \mathbf{z}_\text{w}, \mathbf{z}_\text{c})).
\end{equation}
It should be note that values in the attention map $\mathbf{M}$ are likely to be different from each other, resulting in a fine-grained attention map. Our fine-grained attention map representation method is of major contrast to the previous attention-based methods that learn a coarse attention map, having the same values across elements within the same channel, for example.

\subsection{Feature Calibration}
With the attention tensor estimated in Eq. (\ref{eq:intersection}), we then apply it to the input feature tensor. In particular, we consider computational approaches as follows:
\begin{flalign}
    \mathbf{\hat{X}}=\left\{\begin{array}{ll}
        \mathbf{X}\odot\mathbf{M}&\text{(non-residual)} \\
        \mathbf{X}\oplus(\mathbf{X}\odot\mathbf{M})& \text{(residual)}
    \end{array}\right.
    \label{eq:residual_nonresidual}
\end{flalign}
where $\odot$ and $\oplus$ denote Hadamard product and element-wise summation, respectively.

Fundamentally, although those two approaches employ fine-level attention maps, enabling detailed feature calibration at element-level units, they work in different ways in terms of feature representation learning.
The non-residual approach is to mine and calibrate as many discriminative features as possible by multiplying the input feature tensor with the corresponding attention tensor. Meanwhile, in the residual approach, because of the element-wise sum operation (\ie, a residual operation), the element-wise multiplication part between the input feature tensor and the attention tensor is likely to excite the locations where the input feature tensor $\mathbf{X}$ may have less activations. Hence, the attention module described in Section \ref{sec:attention} is trained to give more attention to locations where the input feature tensor presents relatively small activations. Accordingly, the output attention tensor in $\mathbf{M}$ plays the role of exciting the less activated regions in $\mathbf{X}$ while inhibiting the more activated regions. This interpretable phenomenon is clearly observed from our experimental results in Fig. \ref{fig:activationmap_cam_RFGA} and Fig. \ref{fig:activationmap_cam_RFGA_wo}.

\begin{table*}[t]
\begin{center}
\caption{WSOL performances between our proposed method (RFGA) and comparative methods on CUB-200-2011 (CUB), FGVC Aircraft (Aircraft), and Stanford Dogs (Dogs) datasets. Note that w/ R and w/o R denote with residual connection and without residual connection, and $\tau$ indicates a threshold of activation map. The best performance is indicated in boldface.}
\resizebox{\textwidth}{!}{
\renewcommand{\arraystretch}{1.3}
\Huge{\begin{tabular}{cccccccc ccccccc ccccccc}\\\toprule
\multirow{3}{*}{Method} & \multicolumn{7}{c}{CUB}  &  \multicolumn{7}{c}{Aircraft } & \multicolumn{7}{c}{Dogs }    \\ \cmidrule(lr){2-8}\cmidrule(lr){9-15}\cmidrule(lr){16-22}
 {}                               & \multicolumn{6}{c}{MaxBoxAcc (\%)}   & \multirow{2}{*}{\shortstack{mIoU\\(optimal $\tau$)}}  & \multicolumn{6}{c}{MaxBoxAcc (\%)} & \multirow{2}{*}{\shortstack{mIoU\\(optiaml $\tau$)}} & \multicolumn{6}{c}{MaxBoxAcc (\%)} & \multirow{2}{*}{\shortstack{mIoU\\(optiaml $\tau$)}} \\ \cmidrule(lr){2-7}\cmidrule(lr){9-14}\cmidrule(lr){16-21}
 {}                               & $\delta=0.5$ & $\delta=0.6$ & $\delta=0.7$ & $\delta=0.8$ & $\delta=0.9$ & Avg. &   & $\delta=0.5$ & $\delta=0.6$ & $\delta=0.7$ & $\delta=0.8$ & $\delta=0.9$ & Avg. & & $\delta=0.5$ & $\delta=0.6$ & $\delta=0.7$ & $\delta=0.8$ & $\delta=0.9$ & Avg. &\\\midrule
 { }{ }SENet \cite{hu2018squeeze}        & 69.92 & 39.49 & 15.26 & 4.04 & 0.55 & 25.85 & 0.6598      & 82.81 & 63.67 & 33.30 & 8.88 & 1.95 & 38.12 & 0.7148    & 81.60 & 69.95 & 52.87 & 31.18 & 10.34 & 49.18 & 0.7850 \\
{ }{ }CBAM  \cite{woo2018cbam}           & 68.19 & 39.16 & 15.36 & 3.75 & 0.55 & 25.40 & 0.6538      & 75.67 & 57.88 & 35.04 & 11.97 & 2.16 & 36.54 & 0.7179   & 81.60 & 69.63 & 52.44 & 30.79 & 10.36 & 48.96 & 0.7875 \\
{ }{ }ECA-Net \cite{wang2020eca}         & 69.23 & 39.61 & 15.21 & 3.66 & 0.48 & 25.63 & 0.6603      & 89.41 & 75.46 & 41.31 & 10.44 & 2.16 & 43.75 & 0.7327 & 82.72    & 71.48 & 54.88 & 32.72 & 10.92 & 50.54 & 0.7901 \\\midrule

{ }{ }CAM \cite{Zhou_2016_CVPR}           & 68.54 & 39.59 & 16.00 & 3.92 & 0.54 & 25.71 & 0.6565      & 84.22 & 66.31 & 36.84 & 11.70 & 1.98 & 40.21 & 0.7234      & 81.05  & 69.58 & 53.87 & 31.76 & 10.51 & 49.35 & 0.7884 \\
{ }{ }HaS \cite{singh2017hide}           & 64.43 & 35.88 & 13.74 & 3.40 & 0.54 & 23.59 & 0.6474      & 73.18 & 52.12 & 29.58 & 10.26 & 2.04 & 33.43 & 0.7147    & 78.72 & 67.47 & 51.20 & 30.61 & 10.24 & 47.64 & 0.7833  \\
{ }{ }ACoL \cite{zhang2018adversarial}   & 57.92 & 29.03 & 9.75  & 2.19 & 0.35 & 19.84 & 0.6197      & 51.40 & 28.11 & 12.84 & 5.01 & 1.86 & 19.84 &  0.6259   & 68.88 & 54.55 & 37.84 & 21.26 & 7.55 & 38.01 & 0.7141 \\
{ }{ }ADL  \cite{choe2019attention}      & 62.69 & 33.47 & 12.08 & 3.12 & 0.45 & 22.36 & 0.6402      & 64.36 & 40.41 & 17.94 & 6.00 & 1.92 & 26.12 & 0.6593     & 79.17 & 66.54 & 48.40 & 27.75 & 9.46 & 46.26 & 0.7667 \\
{ }{ }CutMix \cite{yun2019cutmix}        & 72.58 & 45.37 & 19.88 & 5.37 & 0.66 & 28.71 & 0.6724      & 79.75 & 71.83 & 56.95 & 29.04 & 4.86 & 48.48 & 0.7775    & 74.55 & 62.30 & 47.38 & 28.45 & 10.31 & 44.59 & 0.7822 \\\midrule
{ }{ }\textbf{RFGA w/o R} & {71.85} & {42.04} & {17.24} & {4.09} & {0.59} & {27.16} & {0.6635} & \textbf{96.22} & \textbf{87.22} & {53.29} & {15.78} & {2.37} & {50.97} & {0.7600}  & \textbf{84.81} & {74.62} & {59.22} & {36.49} & {12.10} & {53.44} & {0.7613}\\
{ }{ }\textbf{RFGA w/ R (Ours)} & \textbf{75.99} & \textbf{51.19} & \textbf{26.30} & \textbf{8.08} & \textbf{1.19} & \textbf{32.55} & \textbf{0.6970} & {89.11} & {79.15} & \textbf{62.59} & \textbf{35.70} & \textbf{6.48} & \textbf{54.60} & \textbf{0.8101}  & {84.15} & \textbf{74.69} & \textbf{60.55} & \textbf{39.14} & \textbf{13.61} & \textbf{54.42} & \textbf{0.8135}\\\bottomrule
\end{tabular}}}
\label{tab:MaxBoxAcc}
\end{center}
\end{table*}

\section{Experiment}
\subsection{Experimental Setup}
\textbf{Datasets.}
We validated our RFGA over three public datasets for WSOL, \ie, CUB-200-2011 \cite{wah2011caltech}, FGVC Aircraft \cite{maji2013fine}, and Stanford Dogs \cite{KhoslaYaoJayadevaprakashFeiFei_FGVC2011}. First, CUB-200-2011 includes a total of $11,788$ images from $200$ bird categories, which is divided into $5,994$ images for training and $5,794$ images for evaluation. FGVC Aircraft consists of $10,000$ images across $100$ aircraft categories with $3,334$ for training, $3,333$ for a validation, and $3,333$ for testing. Stanford Dogs contains a total of $20,580$ dog samples in $102$ categories, which is composed of $12,000$ training samples and $8,580$ test samples.

\textbf{Competing methods.}
We compared our RFGA with five existing state-of-the-art WSOL methods; \eg, CAM \cite{Zhou_2016_CVPR}, HaS \cite{singh2017hide}, ACoL \cite{zhang2018adversarial}, ADL \cite{choe2019attention}, and CutMix \cite{yun2019cutmix}. Further, in order to see the effectiveness of attention methods in WSOL, we compared to three other context fusion based attention methods; \eg, SENet \cite{hu2018squeeze}, CBAM \cite{woo2018cbam}, and ECA-Net \cite{wang2020eca} as well.

\textbf{Evaluation metric.}
In order for quantitative evaluation, we used MaxBoxAcc \cite{choe2020evaluating} over the IoU thresholds $\delta\in\{0.5,0.6,0.7,0.8,0.9\}$ at the optimal activation map threshold. A threshold of activation map, $\tau$, is set between 0 and 1 at 0.01 intervals. Therefore, we finally obtained the results, measuring various localization performances over threshold $\tau$ for an activation map at various levels of $\delta$.

\subsection{Implementation Details}
We used ResNet-50 \cite{he2016deep} pre-trained with ImageNet data as a backbone network. In order to obtain localization maps, we used feature maps of $1\times1$ convolutional layers, similar to ACoL \cite{zhang2018adversarial}. For the kernel size $k$ in the triple-view attentions, we used $3$, according to the work of \cite{wang2020eca}. The input images of training were resized to $600\times600$ and then we cropped randomly $448\times448$ patches from the resized images. In addition, the input images were flipped horizontally with a probability of $0.5$. Meanwhile, the test images were resized to $448\times448$. We trained our RFGA for a total of $60$ epochs with a mini-batch size of $20$ and an initial learning rate of $0.01$ that was decreased by $0.1$ after every $15$ epochs. Further, we used the stochastic gradient descent optimizer with a momentum of $0.9$. More details for the settings of comparative methods can be found in Supplemantary. We implemented all methods in PyTorch\footnote{\url{https://pytorch.org}} and trained with Titan X GPU. 

\subsection{Experimental Results}
\textbf{Quantitative evaluation.}
Table~\ref{tab:MaxBoxAcc} summarizes the performance of the competing methods at the optimal activation map threshold $\tau$. We observed that our RFGA method outperformed localization performance compared to other competing methods in terms of MaxBoxAcc and Mean Intersection over Union (mIoU) which is the average IoU of all images at optiaml $\tau$. Further, it is noteworthy that our RFGA showed its superiority to all competing methods in all cases including various IoU thresholds over three datasets.

\begin{table}[thp!]
\renewcommand\thetable{2}
\begin{center}
\caption{Classification performances on CUB-200-2011 (CUB), FGVC Aircraft (Aircraft), and Stanford Dogs (Dogs) datasets. Note that w/ R and w/o R denote with residual connection and without residual connection. The best performance is indicated in boldface.}

\scalebox{0.8}{
\begin{tabular}{c ccc}\\\toprule
      \multirow{2}{*}{\textbf{Method}}  & \multicolumn{3}{c}{Classification Top-1 Accuracy (\%)} \\\cmidrule(lr){2-4} 
      {}                                & {CUB}  & {Aircraft} & {Dogs} \\\midrule
        SENet \cite{hu2018squeeze}         & 80.84 & 63.04 & 79.51    \\
        CBAM  \cite{woo2018cbam}           & 80.57 & 58.69 & 79.36    \\
        ECA-Net \cite{wang2020eca}         & 80.31 & 65.62 & 78.55    \\\midrule
        CAM \cite{Zhou_2016_CVPR}          & 81.41 & 63.16 & 79.35    \\
        HaS \cite{singh2017hide}           & 70.54 & 41.97 & 73.15    \\
        ACoL \cite{zhang2018adversarial}   & 61.70 & 11.94 & 47.28    \\
        ADL  \cite{choe2019attention}      & 63.32 & 46.38 & 67.63    \\
        CutMix \cite{yun2019cutmix}        & \textbf{83.55} & 59.59 & \textbf{83.17}    \\\midrule
        \textbf{RFGA w/o R}                & 79.75 & \textbf{65.83} & 76.13    \\
        \textbf{RFGA w/ R (Ours)}                & 76.65 & 52.99 & 68.17 \\\bottomrule
    \end{tabular}}
    \label{tab:Classification}
\end{center}
\end{table}

\begin{figure*}[htp!]
\begin{center}
\includegraphics[width=0.85\linewidth]{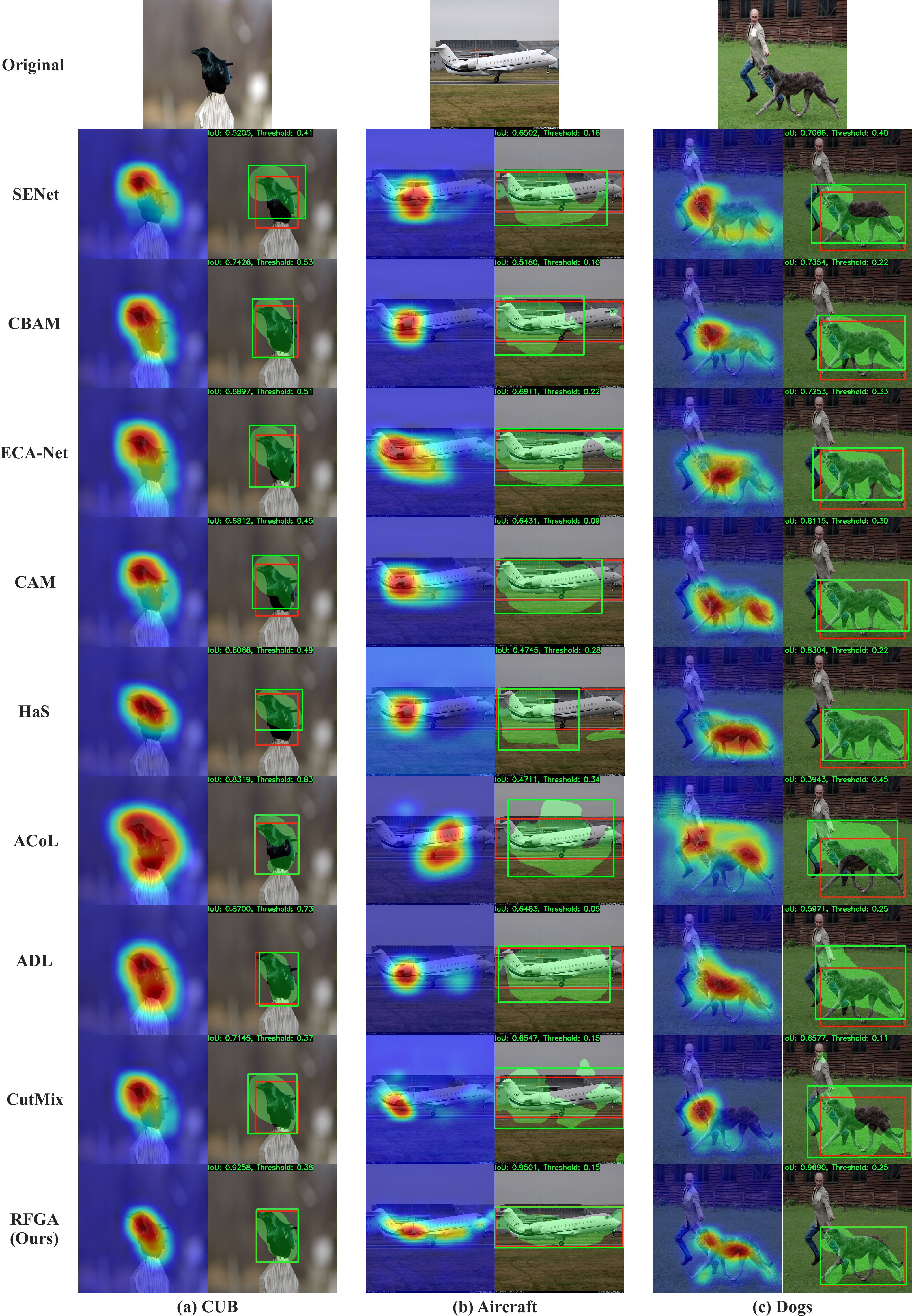}
\end{center}
   \caption{Qualitative comparison between our proposed method (RFGA) and competing methods for WSOL task on (a) CUB-200-2011, (b) FGVC Aircraft, and (c) Stanford Dogs datasets. Our RFGA can generate more exact localization maps by tightly bounding the entire region of the object in an image.}
\label{fig:visualize}
\end{figure*}

\textbf{Qualitative visualization.} We visualized the predicted localization bounding boxes and activation maps for all methods in Fig. \ref{fig:visualize}. Each image presents the localization results at the optimal threshold $\tau$ where the IoU of the bounding box from the activation map achieves maximum value. We observed that RFGA elaborately localized the entire part of an object for CUB-200-2011, FGVC aircraft, and Stanford Dogs datasets. While most competing methods focused on the partial objects or covered in excess of the exact object region, our RFGA tightly bounded the entire and specified region of the object in an image, thereby achieving the best localization performance.

\textbf{Classification.} 
We additionally report the classification accuracy in Table~\ref{tab:Classification} to explore the relation between localization and classification. By following to work of \cite{choe2020evaluation}, we selected the model at the last epoch without regard to validation results. However, we observed that most WSOL methods showed a tendency of achieving the best localization performance in the early stage of training in spite of low classification performance. Consequently, we believe that the classification performance is not correlated with the localization performance, consistent with the work in \cite{choe2020evaluating}.

\begin{figure*}[thp!]
\begin{center}
\includegraphics[width=1.0\linewidth]{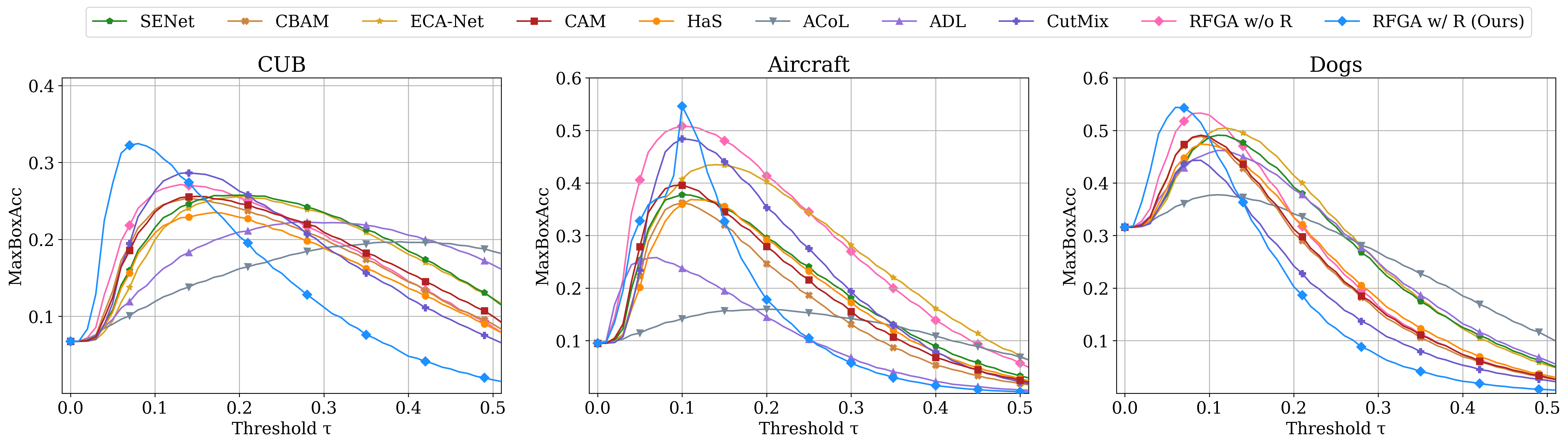}
\end{center}
   \caption{Localization performance by operating an activation map threshold $\tau$ on CUB-200-2011 (left), FGVC Aircraft (middle), and Stanford Dogs (right). Compared to other comparitive methods, our RFGA showed the best MaxBoxAcc at small $\tau$ values for all datasets.}
\label{fig:experiment2}
\end{figure*}
\begin{figure*}[t]
\begin{center}
\includegraphics[width=1.0\linewidth]{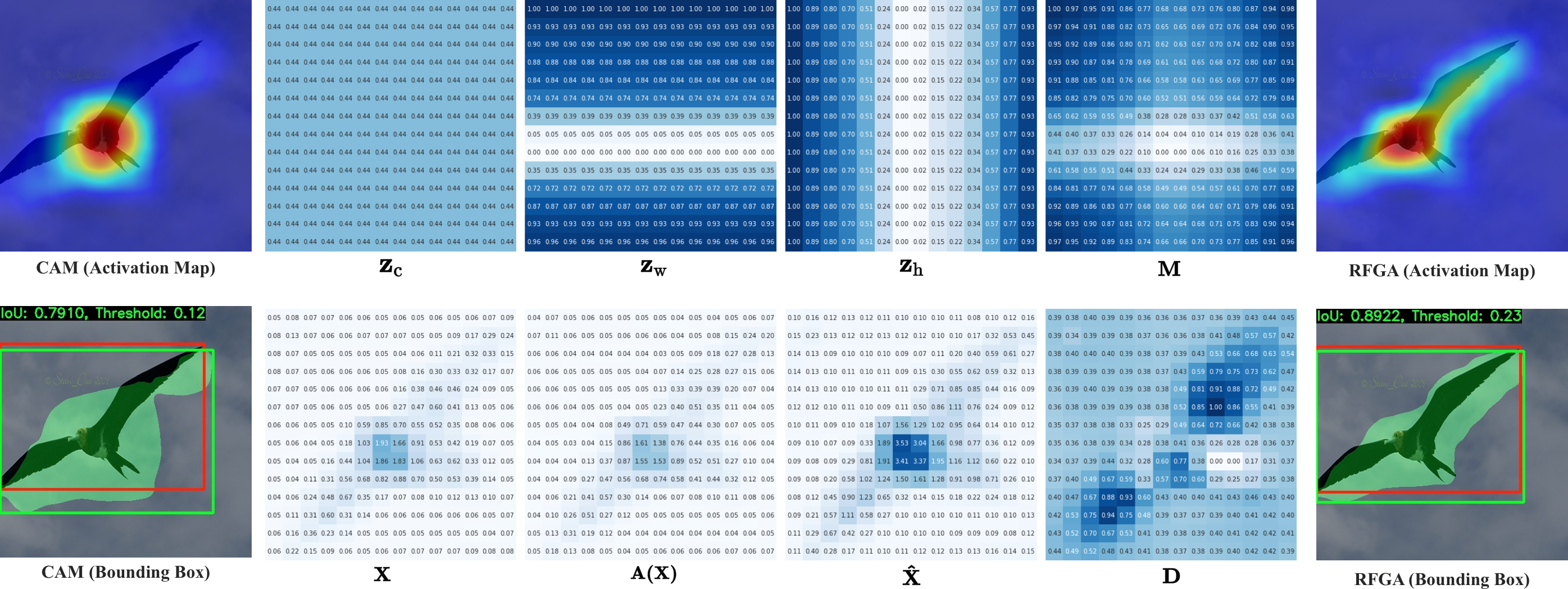}
\end{center}
   \caption{Visualization on activation maps and bounding boxes in CAM~\cite{Zhou_2016_CVPR} and our RFGA for comparison. We plotted all triple-view attention maps in our RFGA by normalizing them in a range between $0$ and $1$ (top). Also, we plotted the normalized difference between $\mathbf{X}$ and $\hat{\mathbf{X}}$, $\mathbf{D}$, to show to which RFGA gives attention.}
\label{fig:activationmap_cam_RFGA}
\end{figure*}

\subsection{Analysis}
\textbf{Hyperparameter analysis.}
We plotted the change of localization performance (MaxBoxAcc \cite{choe2020evaluating}) of all methods by varying the value of $\tau$ in Fig. \ref{fig:experiment2}. Our proposed RFGA showed the best performance at a smaller $\tau$ value than that of other comparative methods. From the results, we could infer that most of the high activation values in our RFGA-based feature tensor were well aligned to the object-related region.

\textbf{Visualization of attention maps.} 
In order to get an insight into the working of our RFGA, we visualized all triple-view attention maps (\ie, $\mathbf{z}_{c}$, $\mathbf{z}_{w}$, $\mathbf{z}_{h}$) as well as the final combined attention map (\ie, $\mathbf{M}$) (top) and the input feature map $\mathbf{X}$, the element-wise product of $\mathbf{X}$ and $\mathbf{M}$ (\ie, $A(\mathbf{X})$), the resulting output feature map $\hat{\mathbf{X}}$ via a residual approach, and the difference between $\mathbf{X}$ and $\hat{\mathbf{X}}$ (bottom) in Fig. \ref{fig:activationmap_cam_RFGA}. We took an average of each attention vector along the channel axis and expanded the averaged vectors, $\mathbf{z}_c$, $\mathbf{z}_w$, and $\mathbf{z}_h$ to a matrix by repetition for a visualization. It should be noted that we normalized each matrix in a range of $[0,1]$. 
Contrary to the activation map of CAM that only focuses on the body of a bird, our RFGA pays additional attention to the wings, resulting in the entire object attention. In accordance with Fig. \ref{fig:visualize}, we validated the effectiveness of our fine-grained calibration of features in WSOL.

\begin{figure*}[t]
\begin{center}
\includegraphics[width=1.0\linewidth]{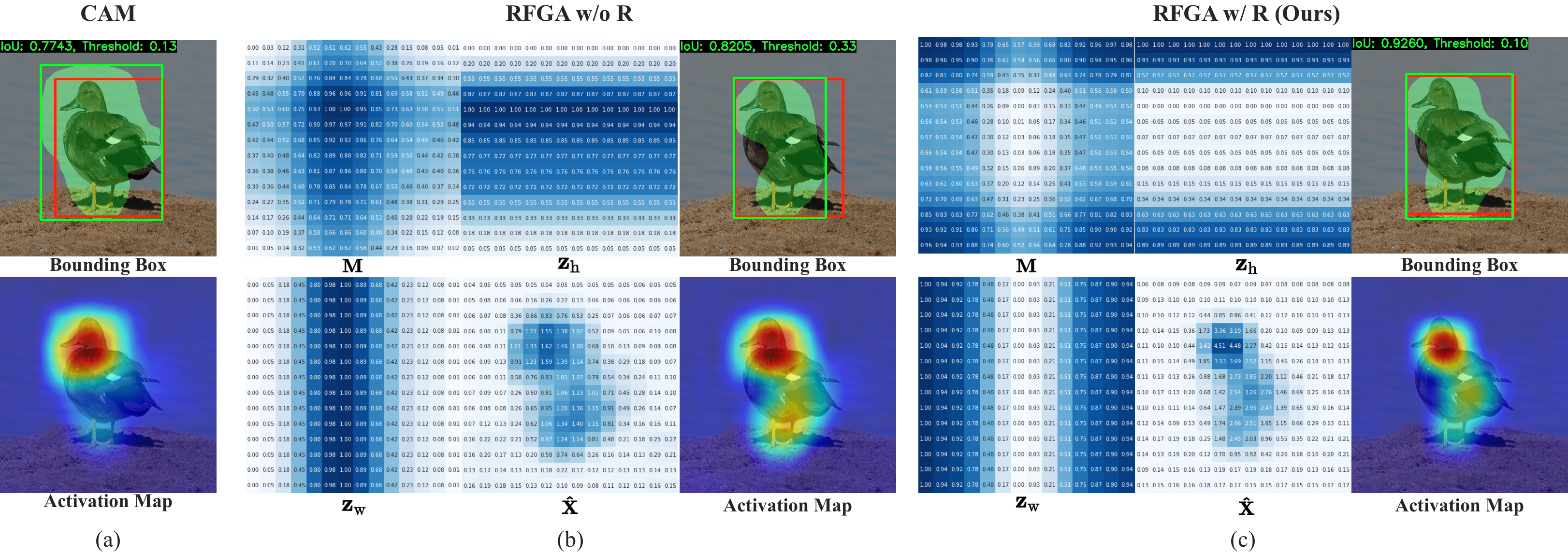}
\end{center}
   \caption{Illustrative comparison of activation maps on (a) CAM \cite{Zhou_2016_CVPR}, (b) RFGA w/o R, and (c) RFGA w/ R. For (b) and (c), we observed the reversed patterns in the attention maps, \ie, $\mathbf{M}$, $\mathbf{z}_\text{h}$, and $\mathbf{z}_\text{w}$, between them. In the w/ R case, the localization performance was much better than the w/o R case and more fine-grained appearance of the object was discovered in the activation map. } 
\label{fig:activationmap_cam_RFGA_wo}
\end{figure*}

\subsection{Ablation Study}
\textbf{Effect of triple-view attention.}
We assumed that our RFGA could localize an object from a variety of fine-grained information. To validate the effectiveness of the fine-grained attention, we conducted ablation studies with respect to each of the attention maps generated from different views, \ie, channel $\mathbf{z}_{c}$, vertical $\mathbf{z}_{h}$, and horizontal $\mathbf{z}_{w}$. We employed only one-view attention out of those three view when training the same architecture, and reported the results in Table~\ref{tab:ablation_dimension}. Our triple-view attentions method clearly outperformed all the ablation cases. Based on those results, we believe that our fine-grained attention map from the triple-view attentions is capable of calibrating features through the complementary relations inherent in the input feature tensor.

\begin{table}[t]
\begin{center}
\caption{Comparison of different attention views and fine-grained attention (denoted as Full) in RFGA. The best performance is highlighted in boldface. Note that we expand the triple-view attentions (channel, height and width) to a fine-grained attention map by an outer sum.} 
\scalebox{0.78}
{\begin{tabular}{@{}cc cc cc @{}}\\\toprule
\multirow{2}{*}{\textbf{\shortstack{Dataset}}}  & \multirow{2}{*}{\textbf{Metric}} & \multicolumn{4}{c}{\textbf{Attention Dimension}}  \\
{} & {} & Channel & Height & Width & Full \\ \midrule
\multirow{2}{*}{\shortstack{CUB }}
{} & MaxBoxAcc          & 31.71 & 6.78 & 31.03 & \textbf{32.55}    \\
{} & mIoU               & 0.6893 & 0.3862  & 0.6865 & \textbf{0.6970}       \\\midrule
\multirow{2}{*}{\shortstack{Aircraft}}
{} & MaxBoxAcc          & 52.65 & 21.55 & 45.53 & \textbf{54.60}     \\
{} & mIoU               & 0.7866 & 0.6120 & 0.7556 & \textbf{0.8101} \\\midrule
\multirow{2}{*}{\shortstack{Dogs }}
{} & MaxBoxAcc          & 52.81    & 30.21   & 53.13 & \textbf{54.42}        \\
{} & mIoU               & 0.8001   & 0.5710  & 0.8018 & \textbf{0.8135}    \\\bottomrule
\end{tabular}
\label{tab:ablation_dimension}
}
\end{center}
\end{table}

\textbf{Effect of residual connection.} 
In order to investigate residual connection effect, we compared the residual approach and the non-residual approach in Eq. (\ref{eq:residual_nonresidual}) in terms of the localization task in Table~\ref{tab:MaxBoxAcc} and the classification task in Table~\ref{tab:Classification}. We also visualized their respective attention maps in Fig. \ref{fig:activationmap_cam_RFGA_wo}. The residual approach generated the attention maps that focused on both the most and the less discriminative regions of an object. Meanwhile, for the non-residual approach, their attention maps showed the opposite patterns to those of the residual-based maps. 
Based on the understanding of a residual operation, note that $\mathbf{X}\oplus A(\mathbf{X})$ leads the function $A(\mathbf{X})$ to learn some amount of information that the input feature tensor $\mathbf{X}$ may have missed or less emphasized. From the viewpoint of attention map generation, the role of $A(\mathbf{X})$ can be interpreted as to inhibit the regions of high activation values (as those are already well represented in $\mathbf{X}$) and to excite the less activated regions where the target task-related information is inherent. Here, the inhibition effect can be related to those of the specially-designed module in ACoL\cite{zhang2018adversarial} and ADL\cite{choe2019attention} that erase the discriminative features.

\section{Conclusion}
In this paper, we proposed a novel residual fine-grained attention module to localize an object accurately. Our proposed RFGA consisted of three components; (i) the triple-view attentions, (ii) expansion of the attentions to a high resolution, and (iii) calibration of the feature. Notably, our proposed method does not require a hyperparameter such as a drop rate for masking discriminative regions. Based on the evaluation with the metrics of mIOU and MacBoxAcc \cite{choe2020evaluating} over three datasets, our proposed method achieved the highest performance. In our exhaustive ablation studies, we presented the validity of all the three components and also interpreted the inner working of the feature calibration formulated by a residual operation. It is noteworthy that because our proposed RFGA is plugged in between the last convolution layer and a classifier, it is applicable to other CNN architectures without modifying the original network architecture. In that sense, it would be our forthcoming research issue to more generalize its application to multi-object localization.

\textbf{Acknowledgement.} This work was supported by Institute of Information $\&$ communications Technology Planning $\&$ Evaluation (IITP) grant funded by the Korea government (MSIT) (No. 2019-0-00079 , Artificial Intelligence Graduate School Program(Korea University)).
{\small
\bibliographystyle{ieee_fullname}
\bibliography{egbib}
}


\end{document}